\documentclass{article}

\usepackage{arxiv}
\usepackage{graphicx}
\usepackage{makecell}

\usepackage[utf8]{inputenc} 
\usepackage[T1]{fontenc}    
\usepackage{hyperref}       
\usepackage{url}            
\usepackage{booktabs}       
\usepackage{amsfonts}       
\usepackage{nicefrac}       
\usepackage{microtype}      
\usepackage{xcolor}         
\usepackage{color,soulutf8}
\usepackage{multirow}
\usepackage{dirtytalk}

\title{Otter-Knowledge: benchmarks of multimodal knowledge graph representation learning from different sources for drug discovery
}

\author{%
  Hoang Thanh Lam, Marco Luca Sbodio, Marcos Mart\'inez Galindo, Mykhaylo Zayats, \\ \textbf{Ra\'ul Fern\'andez-D\'iaz, V\'ictor Valls, Gabriele Picco, Cesar Berrospi Ramis, Vanessa L\'opez }\\
  IBM Research Europe \\ Dublin Lab, Ireland and Zurich Lab, Switzerland\\
  \small\texttt{t.l.hoang@ie.ibm.com, marco.sbodio@ie.ibm.com, Marcos.Martinez.Galindo@ibm.com}\\ 
  \small\texttt{mykhaylo.zayats1@ibm.com, raulfd@ibm.com, victor.valls@ibm.com} \\
  \small\texttt{gabriele.picco@ibm.com, ceb@zurich.ibm.com and vanlopez@ie.ibm.com} 
}

\begin{document}

\maketitle

\begin{abstract}
Recent research on predicting the binding affinity between drug molecules and proteins use representations learned, through unsupervised learning techniques, from large databases of molecule SMILES and protein sequences. While these representations have significantly enhanced the predictions, they are usually based on a limited set of modalities, and they do not exploit available knowledge about existing relations among molecules and proteins. In this study, we demonstrate that by incorporating knowledge graphs from diverse sources and modalities into the sequences or SMILES representation, we can further enrich the representation and achieve state-of-the-art results for drug-target binding affinity prediction in the established Therapeutic Data Commons (TDC) benchmarks. We release a set of multimodal knowledge graphs, integrating data from seven public data sources, and containing over 30 million triples. Our intention is to foster additional research to explore how multimodal knowledge enhanced protein/molecule embeddings can improve prediction tasks, including prediction of binding affinity. We also release some pretrained models learned from our multimodal knowledge graphs, along with source code for running standard benchmark tasks for prediction of biding affinity.
\end{abstract}

\section{Introduction}
\label{sec:introduction}
Developing a concise representation of proteins and small molecules is a crucial task in AI-based drug discovery. Recent studies \cite{NEURIPS2019_37f65c06, rives2019biological} have focused on utilizing large databases of protein sequences or molecules for  self-supervised representation learning. These representations are then fine-tuned using limited labeled data for tasks like predicting the binding affinity between drugs and targets. In the field of protein representation learning, \cite{zhang2022ontoprotein} and \cite{zhou2023protein} have demonstrated that enhancing protein representations with additional information from knowledge graphs (KGs), such as comprehensive textual data from the gene ontology (GO) \cite{Ashburner_2000}, can enhance the performance of pre-trained representations on various protein properties and protein-protein interaction tasks. Despite promising initial findings, important research directions remain unexplored.
 
 
Firstly, it is important to investigate if  joint robust representations of proteins, molecules and other associated entities (such as diseases) may improve downstream prediction tasks. A joint representation of each of these entities should take into account not only the intrinsic properties of the entity itself, but also its relations with other entities. In contrast to previous studies \mbox{\cite{zhang2022ontoprotein, zhou2023protein}}, concentrating on protein sequences, GO terms, and text descriptions, our work encompasses a wider array of modalities—text, numbers, protein sequences, SMILES, categories, and entities like drugs, diseases, and pathways.

 
 Secondly, previous studies \mbox{\cite{zhang2022ontoprotein, zhou2023protein}} assumed that the graphs or datasets used for training were carefully combined into a single source. However, in our research, we deal with different KGs built using several datasets from various sources. It's worth highlighting that merging these datasets into a single source is complex due to the difficulty of aligning their structures automatically. For example, in the STITCH knowledge graph \mbox{\cite{szklarczyk2016stitch}}, the connection labeled ``interaction with", denoting the relationship between chemicals and proteins, is established by analyzing their co-occurrence within a Pubmed abstract. This form of association might bear resemblance to the ``target of" relation found in the Uniprot knowledge graph, however confirming their equivalence presents a challenging task because co-orcurrence in a Pubmed abstract does not mean a protein is a target of a drug.
 Moreover, creating a comprehensive graph by combining multiple ones demands significant computational resources. To address these issues, we propose an approach based on ensemble methods, and we show that it can effectively capture information from a subset of our multimodal knowledge graphs without merging them.

Finally, while there are datasets for studying the problem of drugs-proteins interactions, there are no publicly available multimodal knowledge graphs for this purpose. We aim to provide the research community with such graphs, by selectively integrating knowledge from existing datasets.


The contributions of this work are summarised as follows:
 \begin{itemize}

     \item We release to the research community a set of multimodal knowledge graphs, and a set of  models pretrained on those graphs using graph neural networks to derive protein and drug representations (\url{https://github.com/IBM/otter-knowledge}). We hope that this will foster further research in the area.

     \item We provide experimental results showing that a multimodal knowledge enhanced representation surpasses the state of the art for predicting drugs-proteins interactions, even for challenging discovery scenario, where majority of entities in the test data (e.g., molecules) are unseen in the training data and might only have one available modality (e.g, its SMILES). Previous work focused on enhancing Protein Language Models (PLMs) with textual descriptions, and did not exploit KGs that capture heterogenous relations between entities and multimodal entity attributes (including molecules SMILES, proteins sequences, textual, numerical and categorical attributes).

    \item We tackle the problem of learning from partially connected multimodal knowledge graphs via ensemble methods, yielding promising outcomes. We also study how our approach remains robust across diverse pretraining objectives, such as regression and link prediction.
 \end{itemize}

\section{Multimodal knowledge representation learning}
\begin{figure}[h]
\centering
\includegraphics[width=\textwidth]{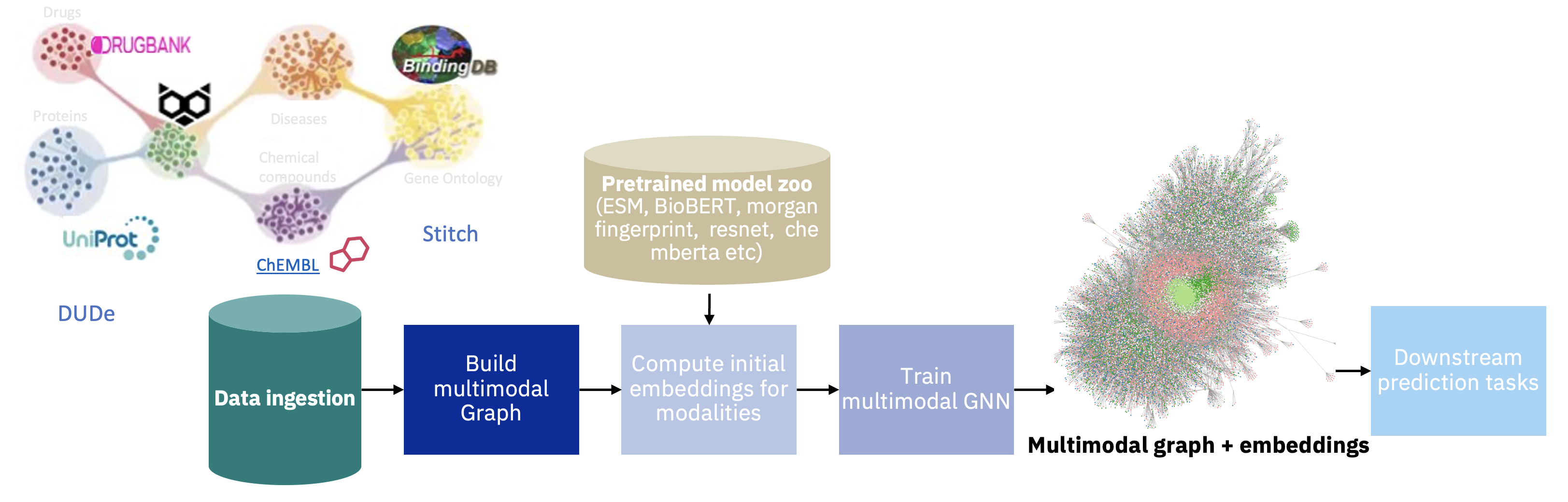}
\caption{ \emph{Otter-Knowledge} workflow.}
\label{fig:workflow}
\end{figure}
 The diagram in Figure \ref{fig:workflow} illustrates the overall process of our system named \emph{Otter-Knowledge}. This process involves constructing multimodal knowledge graphs from diverse sources, generating initial embeddings for each modality using pretrained models available in the model zoo, and subsequently improving the embeddings by incorporating information from the knowledge graphs through the utilization of graph neural networks (GNN). We discuss each component of the given system in the following subsections.



\subsection{Multimodal Knowledge graph construction}

A Multimodal Knowledge Graph (MKG) is a directed labeled graph where labels for nodes and edges have well-defined meanings, and each node has a modality, a particular mode that qualifies its type (text, image, protein, molecule, etc.). We consider two node subsets: \textit{entity nodes} (or entities), which corresponds to concepts in the knowledge graph (for example protein, or molecule), and \textit{attribute nodes} (or attributes), which represent qualifying attributes of an entity (for example the mass of a molecule, or the description of a protein). We refer to an edge that connects an entity to an attribute as \textit{data property}, and an edge that connects two entities as \textit{object property}. Each node in the graph has a unique identifier, and a unique modality (specified as a string).

We developed a framework for automating the construction of a multimodal knowledge graph by extracting and fusing data from a variety of sources, including text delimited files, JSON, and proprietary data sources \cite{https://doi.org/10.1002/ail2.20}. The framework takes as input a schema file (specified in JSON), which declaratively describes how to build the desired graph from a set of data sources; a user may extend predefined schema files used to create our multimodal KGs (see our github repository), or define custom ones.


The framework that builds the MKG ensures that each triple is unique, and it automatically merges entities having the same unique identifier, but whose data is extracted from different data sources. It is also possible to use the special relation \texttt{sameAs} \footnote{We borrow the semantic of \texttt{owl:sameAs} - see \url{https://www.w3.org/TR/owl-ref/\#sameAs-def}} to indicate that two entities having different unique identifiers are to be considered as the same entity. The \texttt{sameAs} relation is useful when creating a MKG from multiple partially overlapping data sources; when the graph is built.
Additionally, it is possible to build an MKG incrementally, by merging two or more graphs built using different schemas; The merge operation automatically combines entities with matching unique identifiers or distinct attributes (e.g., proteins with the same sequence).

The framework builds the graph in memory, but can also provide support for building the graph using a database on disk; the graph triples can also be serialised using GML \footnote{\url{https://en.wikipedia.org/wiki/Graph_Modelling_Language}} or any RDF \footnote{\url{https://www.w3.org/RDF/}} serialization formats. Finally, the framework offers scalable parallel and GPU-based computation of initial node embeddings, tailored to each node's modality.

\subsection{Computing initial embeddings}
\label{subsec:initial_embeddings}
As explained before, the MKG contains nodes representing entities and nodes representing attributes of those entities. In the MKG, each node has a modality assigned, e.g: entity nodes can have a modality \emph{Protein} or \emph{Drug}, nodes containing text could have a modality \emph{text} or protein sequence. We assign a model for each one of the modalities in our graphs, as specified by the user in the \emph{schema}. The models, referred to as \emph{handlers}, are capable of preprocessing the values in the nodes and computing their initial embeddings. Our framework allows to easily retrieve all the nodes in the graph with the same modality to efficiently compute the initial embeddings with the assigned \emph{handler}, facilitating parallelization, GPU utilization, batching, and avoiding the need to load different models in memory simultaneously.

Handlers are exclusively assigned to nodes representing attributes. For each modality, only one \emph{handler} is being used, although it is possible to change the \emph{handler} assigned. For instance, for SMILES it is possible to use \emph{morgan-fingerprint} or \emph{MolFormer}. The comparison between different models is not within the scope of this work. These are the \emph{handlers} that we have used for computing the initial embeddings of the graph:
\begin{itemize}
    \item \emph{morgan-fingerprint} We use the Morgan fingerprint in RDKit\footnote{\url{https://www.rdkit.org/docs/index.html}} for processing the SMILES. 
     We employ a shape of 2048 (nBits) and a radius of 2 for the Morgan fingerprint. If RDKit fails to generate the fingerprint, we supply an embedding of the same shape filled with zeros.
    \item \emph{MolFormer} \cite{molformer} is a large-scale chemical language model designed with the intention of learning a model trained on small molecules which are represented as SMILES strings. \emph{MolFormer} leverges Masked Language Modeling and employs a linear attention Transformer combined with rotary embeddings.
    \item \emph{protein-sequence-mean} For the protein sequences, we use the mean token embeddings extracted from the 33rd layer of the \emph{esm1b\_t33\_650M\_UR50S} model   \cite{rives2019biological}. All sequences are truncted to 1022 amino acids.
    \item \emph{text} For the textual values, we use the `sentence-transformers/paraphrase-albert-small-v2` \cite{reimers-2019-sentence-bert} \footnote{https://huggingface.co/sentence-transformers/paraphrase-albert-small-v2}.
    \item \emph{number} In the case of numbers, we do not use any model to get initial embeddings. We convert the numerical value to a torch tensor and use it as embedding. 
    \item \emph{categorical, protein and molecule entities} modalities do not possess handlers, resulting in the absence of an initial embedding. Initial embeddings start as zeros and improve through inductive GNN training, based on the neighborhood connections to their data properties and related entities. To illustrate this, an effective representation for an entity Protein is learned through iterative aggregation of information from adjacent nodes representing multimodal attributes (such as its sequence, other textual attributes like label, family, function) initialised with their respective handlers and interactions with other entity nodes (proteins, molecules or categories), provided such modalities and relations exist.

\end{itemize}
Using distinct models lets us leverage existing competitive pretrained language models for meaningful modality representations (e.g., gauging embedding similarity). This aids enhancing generalization, the GNN can exploit cross-modal links through KG geometry and it reduces the need for a larger training dataset to generalise to unseen entities, which would have been necessary if we had opted for a single unified model setup for all modalities.

\subsection{Pretraining with inductive R-GNN}
\label{sec:pretraining}
\paragraph{GNN architecture} To improve representations of MKG entities we assimilate initial embeddings with the information about the MKG structure, i.e., connection patterns between graph nodes. To this end we train GNN \cite{ZHOU2020graph} that propagates initial embeddings of attribute nodes through a set of layers that transform input embedding according to the entity data and object properties. Our GNN is inductive since it may compute entities embeddings only from their available neighbours and the corresponding initial embeddings. This allows embeddings computation of unseen nodes during the training.

The architecture of GNN adheres to a standard design and consists of two main blocks: encoder and decoder. For the encoder, we first define a projection layer, which consists of a set of linear transformations for each attribute node modality and projects those into a common dimensionality. Then, we apply several multi-relational graph convolutional layers (R-GCN)~\cite{Schlichtkrull2018Modeling}, which distinguish between different types of edges connecting source and target nodes by having a set of trainable parameters for each edge type. For the decoder, we consider a link prediction task, which consists of a scoring function that maps each triple of source and target nodes, along with their corresponding edge, to a scalar number defined over interval $[0;1]$. 

\paragraph{Learning objective} For link prediction, we evaluate three common scoring functions: DistMult, TransE, and a Binary Classifier. We compare their scores against actual labels using negative log likelihood loss. Additionally, for numerical attributes, we include a regression objective to minimize root mean square error (MSE) of predicted numerical properties. The learning process combines the link prediction and regression objectives into a single function. GNN hyperparameters values are detailed in the Appendix.

\paragraph{Negative sampling} To train the GNN for link prediction we need to provide both positive and negative examples. While positive examples come from the MKG itself, negative triples are generated synthetically. To achieve this, for each relation type we extract a set of admissible source and target nodes, subsequently, we randomly sample sources and targets from the corresponding admissible sets. We use an equal ratio between positive and negative links.

\paragraph{Scaling the GNN training} Due to the integration of data from various sources, the size of the integrated data can become significantly large. For instance, the combination of Uniprot, ChemBL, and BindingDB requires over 200GB of CPU memory for training, which renders GNN training on such a large graph using limited GPU memory highly inefficient.
To address this, we employ a graph auto-scaling approach (GAS) described in \cite{gas}. This method divides the graph into smaller partitions using Metis\footnote{\url{https://github.com/KarypisLab/METIS}}. GAS performs training on each partition separately, so it is able scale to arbitrarily large graphs. To avoid information loss due to the connection between partitions, it keeps node embeddings of previous training step in CPU memory thus saving GPU memory and uses this historical embeddings to update the nodes inside a partition. Please refer to Appendix for details about the hyper-parameter settings of Metis.

\paragraph{Information flow control}
One crucial aspect of pretraining the GNN involves addressing the disparity between the data accessible during pretraining and the data accessible during subsequent tasks. Specifically, during pretraining, there are numerous attributes associated with proteins or drugs, whereas during downstream fine-tuning, only protein sequences and SMILES are available. Consequently, during pretraining, we explore two scenarios: one which controls the information propagated to the Drug/Protein entities and one without such control. Our experiments cover both cases to shed light on how constraining information flow during pretraining affects subsequent tasks. 

\paragraph{Noisy link prediction}
An additional significant consideration is the presence of noisy links within the up-stream data and how they affect the downstream tasks. To investigate the potential impact on these tasks, we manually handpick a subset of links from each database that are relevant to drug discovery (see details in the Appendix). We then compare the outcomes when training the GNN using only these restricted links versus using all possible links that are potentially noisy with respect to the downstream tasks.

\section{Pretraining datasets collections}
Table \ref{tab:pretraining} summarizes the multimodal KGs used for pretraining our models; they are based on existing KGs and datasets, and in this section we discuss the data preprocessing steps that we applied to the original KGs and datasets. We open-sourced the preprocessed KGs: we refer to the appendices and our github repository for further details.

\begin{table}[]
\small
\centering
\begin{tabular}{llllll}
\hline
\textbf{Datasets} & \textbf{\# triples} & \textbf{Entities} & \textbf{Modalities} & \textbf{Data license} & \textbf{Released} \\ \hline
Otter-UBC           &   6,207,654      &   Proteins/Drugs      &  \makecell{sequences, SMILES, \\text, number}      & Open     & Yes   \\
Otter-PrimeKG           & 12,757,257     & Proteins/Drugs/Diseases     & \makecell{sequences, SMILES, \\text}      & Open     & Yes      \\
Otter-DUDe              & 40,216              & Proteins/Drugs     & sequences, SMILES    & Open            & Yes                \\
Otter-STITCH            & 12,621,873          & Proteins/Chemicals & sequences, SMILES    & Open                  & Yes               \\ \hline
\end{tabular}
\caption{\label{tab:pretraining} Summary of the KGs for pretraining models with the number of triples used for GNN training. More details on the modalities, entities and data properties are discussed in the Appendices.}
\end{table}


\paragraph{Uniprot} \cite{uniprot} comprises 573,227 SwissProt proteins (from curated UniProt subset). The Otter-UBC KG combines all UniProt proteins with diverse attributes, including sequence (567,483 entries), full name, organism, protein family, function, catalytic activity, pathways and length. The KG also features 38,665 \emph{target\_of} edges linking UniProt IDs to ChEMBL and Drugbank IDs, along with 196,133 interactants connecting UniProt protein IDs.

\paragraph{BindingDB} \cite{liu2007bindingdb} consists of 2,656,221 data points involving 1.2 million compounds and 9,000 targets. Rather than using the affinity score, we create triples for every drug-protein pair. To avoid data redundancy, we remove duplicated triples from the TDC DTI dataset. This results in a final set of 2,232,392 triples included in Otter-UBC KG.

\paragraph{ChEMBL} \cite{chembl} consists of drug-like bioactive molecules, in Otter-UBC KG we included 10,261 ChEMBL molecule IDs with their corresponding SMILES downloaded from OpenTargets \mbox{\cite{opentargets}}, from which 7,610 have a \textit{sameAs} link to Drugbank ID molecules. 

\paragraph{Drugbank} \cite{drugbank} consists of detailed chemical data on 9,749 drugs (such as SMILES, description, indication, mechanism of action, organism, average mass, toxicity and other calculated and experimental properties), drug pathways and 1,301,422 drug interactions. Due to licenses restrictions, in Otter-UBC we only release drugbank IDs linked from other sources, such as Uniprot and ChemBL. 

\paragraph{DUDe} \cite{dude} comprises a collection of 22,886 active compounds and their corresponding affinities towards 102 targets. For our study, we utilized a preprocessed version of the DUDe \cite{sledzieski2022adapting}, which includes 1,452,568 instances of drug-target interactions. To prevent any data leakage, we eliminated the negative interactions and the overlapping triples with the TDC DTI dataset. This yielded 40,216 pairs of drug-target interactions.

\paragraph{PrimeKG} (the Precision Medicine Knowledge Graph) \cite{chandak2022building} integrates 20 biomedical resources, it describes 17,080 diseases with 4 million relationships. PrimeKG includes nodes describing Gene/Proteins (29,786) and Drugs (7,957 nodes). Our multimodal KG derived from PrimeKG contains 13 types of modalities and 12,757,300 edges. These edges consist of 154,130 data properties and 12,603,170 object properties. Among them, 642,150 edges represent protein interactions, 25,653 edges represent drug-protein interactions, and 2,672,628 edges represent drug interactions. Our multimodal KG extends the original PrimeKG graph by adding SMILES for listed drugs, and sequences for listed Uniprot proteins.

\paragraph{STITCH} (Search Tool for Interacting Chemicals) \cite{szklarczyk2016stitch} is a database of known and predicted interactions between chemicals represented by SMILES strings and proteins whose sequences are taken from STRING database \cite{szklarczyk2023steing}. 
For the MKG curation we filtered only the interaction with highest confidence, i.e., the one which is higher 0.9. This resulted into 10,717,791 triples for 17,572 different chemicals and 1,886,496 different proteins. Furthermore, the graph is split into 5 subgraphs of roughly similar size so the GNN is trained sequentially on each of them, building on previous subgraph's model.

\section{Experiments}

Here we summarise the main results from some of our experiments demonstrating the benefits of our multimodal KGs. Further details are available in the appendices.

\subsection{Benchmark tasks and downstream models}

\paragraph{Downstream benchmarks} 
To evaluate the performance of the proposed approach on drug-target binding affinity prediction task we use three datasets: DTI DG, DAVIS and KIBA, which are available from the TDC \mbox{\cite{Huang2022}} benchmark.
The DTI DG dataset features a leaderboard with the state-of-the-art metrics reported for different methods. The dataset's temporal split, based on patent application dates, makes this dataset suitable for evaluating method generalization. In contrast, the DAVIS and KIBA datasets employ random splits, including two additional splits based on target or drug. These latter splits assess learning methods with new drugs/proteins.

\paragraph{Downstream models}
TDC framework adapts the DeepDTA approach \mbox{\cite{deepdta}} for drug protein binding affinity prediction model referred to as the downstream model. It consists of 3 blocks: i) convolution block which performs a separate feature learning of input drugs and proteins, ii) concatenation which concatenates drugs and proteins features, and iii) transformation which executes the final transformation. We observed that the given architecture is not optimal when both drug and protein embeddings are already given as input. For this, we compared the original downstream model against a simplified variant that employs only concatenation and transformation blocks. When employing the original downstream model with ESM embeddings for proteins and Morgan fingerprints for molecules SMILES as inputs, we obtained inferior results (0.538, the baseline used in Table \mbox{\ref{tab:leaderboard}}) compared to using the simplified downstream model (0.569) on the DTI DG dataset. This motivated us to created a new model architecture for evaluating the pretrained embeddings. In addition to the network that concatenates and transforms the ESM and Morgan fingerprints, we added a parallel network that concatenates and transforms the GNN embeddings. The outputs from both networks are summed up and returned as the final binding affinity prediction. We note that the new downstream model uses the same mean squared error loss as the original downstream model (for details about the network size refer to the Appendices).

\paragraph{Ensemble learning} The pretrained representation is sensitive to various factors, such as the chosen objectives for the GNN and the specific graphs used for training. Additionally, combining all the datasets into a single large graph requires substantial computational resources and poses challenges in aligning different databases. In this study, we propose a simple approach that involves combining pretrained embeddings obtained from different settings and KGs. This allows us to train multiple GNN models simultaneously without the need to merge all KGs into a single location, resulting in savings in computational resources and effort required for data and schema alignment. We create a linear ensemble by assigning equal weights to individual downstream models, each trained with separate pretrained embeddings given by individual GNN models.

\begin{table}[]
\centering
\begin{tabular}{llll}
\hline
Datasets      & \textbf{DTI DG} & \textbf{DAVIS}     & \textbf{KIBA}      \\ \hline
\# triples    & 232460          & 27621              & 118036             \\
\# drugs      & 140745          & 68                 & 2068               \\
\# proteins   & 569             & 379                & 299                \\
Type of splits & Temporal        & Random/Drug/Target & Random/Drug/Target \\ \hline
\end{tabular}
\label{tab:data statistics}
\caption{TDC benchmark datasets and statistics.}
\end{table}

\begin{table}[]
\small
\begin{tabular}{lllllllll}
\hline
                            & \textbf{Downstream}                                                                                                   & \textbf{DTI DG}            & \multicolumn{3}{c}{\textbf{DAVIS}}                                                                                       & \multicolumn{3}{c}{\textbf{KIBA}}                                                                   \\ \cline{3-9} 
\textbf{Upstream} & Splits                                                                                                                         & \multicolumn{1}{l|}{Temporal}                        & Random                          & Target                          & \multicolumn{1}{l|}{Drug}                            & Random                          & Target                          & Drug                            \\ \hline
                            & Leaderboard\footnote{\url{https://tdcommons.ai/benchmark/dti_dg_group/bindingdb_patent/}}  & \multicolumn{1}{l|}{0.538}                           & NA                              & NA                              & \multicolumn{1}{l|}{NA}                              & NA                              & NA                              & NA                              \\
                            & Baseline                                                                                                         & \multicolumn{1}{l|}{0.569}                           & 0.805                           & 0.554                           & \multicolumn{1}{l|}{\textbf{0.264}} & 0.852                           & 0.630                           & 0.576                           \\ \hline
\multirow{3}{*}{Otter-UBC}        & Otter DistMult       & \multicolumn{1}{l|}{0.578}  & 0.808  & 0.572   & 
                            \multicolumn{1}{l|}{0.152}                           & 0.859  & 0.627   & 0.593                                  \\
                            &  Otter TransE       & \multicolumn{1}{l|}{0.577}  &  0.807  & 0.571   & 
                            \multicolumn{1}{l|}{0.130}                          & 0.858   & 0.644   & 0.583                                \\
                            & Otter Classifier    & \multicolumn{1}{l|}{0.580}  & 0.810   & 0.573   & 
                            \multicolumn{1}{l|}{0.104}                          & 0.861   & 0.631   &  0.616                                 \\ \hline
\multirow{3}{*}{Otter-DUDe}       &Otter DistMult                                                                                                                              & \multicolumn{1}{l|}{0.577}                                &  0.805                               &   0.573                              & \multicolumn{1}{l|}{0.132}                                &  0.857                               &  0.650                               & 0.607                                \\
                            &Otter TransE                                                                                                                             & \multicolumn{1}{l|}{0.576}                                & 0.807                               &  0.570                              & \multicolumn{1}{l|}{0.170}                                &   0.858                              &       0.653                          &  0.604                               \\
                            &Otter Classifier                                                                                                                                & \multicolumn{1}{l|}{0.579}                                &  0.808                                &  0.574                                & \multicolumn{1}{l|}{0.167}                                &   0.860                              &   0.641                              & 0.630                                \\ \hline
\multirow{3}{*}{Otter-PrimeKG}    & Otter DistMult                                                                                                                               & \multicolumn{1}{l|}{0.575}                                & 0.806                                & 0.571                                & \multicolumn{1}{l|}{0.162}                                & 0.858                                & 0.611                                & 0.617                                \\
                            & Otter TransE                                                                                                                              & \multicolumn{1}{l|}{0.573}                                & 0.807                                & 0.568                                & \multicolumn{1}{l|}{0.186}                                & 0.858                                & 0.642                                & 0.607                                \\
                            & Otter Classifier                                                                                                                            & \multicolumn{1}{l|}{0.576}                                & 0.813                                & 0.576                                & \multicolumn{1}{l|}{0.133}                                & 0.861                                & 0.630                                & 0.635                                \\ \hline
\multirow{3}{*}{Otter-STITCH}     & Otter DistMult                                                                                                                               & \multicolumn{1}{l|}{0.575}                                & 0.808                                & 0.573                                & \multicolumn{1}{l|}{0.138}                                & 0.859                                & 0.615                                & 0.603                                \\
                            &Otter TransE                                                                                                                               & \multicolumn{1}{l|}{0.578}                                & 0.814                                & 0.572                                & \multicolumn{1}{l|}{0.119}                                & 0.859                                & 0.636                                & 0.635                                \\
                            &Otter Classifier                                                                                                                                & \multicolumn{1}{l|}{0.576}                                & 0.804                                & 0.571                                & \multicolumn{1}{l|}{0.156}                                & 0.856                                & 0.627                                & 0.585                                \\ \hline
                            & Otter Ensemble                                                                                                       & \multicolumn{1}{l|}{\textbf{0.588}} & \textbf{0.839} & \textbf{0.578} & \multicolumn{1}{l|}{0.168}                            & \textbf{0.886} & \textbf{0.678} & \textbf{0.638} \\ \hline
\end{tabular}
\caption{\label{tab:leaderboard} Results of knowledge enhanced representation on three standard drug-target binding affinity prediction benchmarks datasets with different splits. The evaluation metrics is Pearson correlation (higher is better). We reported the results concerning pretraining on separate upstream datasets and the ensemble of these models.}
\end{table}

\begin{table}[]
\begin{tabular}{llllllll}
\hline
Datasets (Otter-UBC)                  & \textbf{DTI DG} & \multicolumn{3}{c}{\textbf{DAVIS}}               & \multicolumn{3}{c}{\textbf{KIBA}}                \\ \cline{2-8} 
Splits                     & Temporal        & Random         & Target         & Drug           & Random         & Target         & Drug           \\ \hline

\hline
Otter DistMult (C) & 0.575 & 0.809	& 0.571	& 0.126	& 0.861	& \textbf{0.643}	& 0.617 \\
Otter TransE (C) & 0.576 & 	0.809 &	0.570 & \textbf{0.157} &	0.858 &	0.632	& 0.585 \\
Otter Classifier (C) & 0.578 & 	0.814	& \textbf{0.577}	& 0.097	& 0.861	& 0.633 & 	0.631 \\
\hline
Otter DistMult (N+C)  & 0.578 &	0.809 & 0.574 & 0.105 & 0.862 & \textbf{0.643} & 	0.615 \\
Otter TransE (N+C)   & 0.579 & 0.809 & 0.573 & 0.108 & 0.857 & 0.633	& 0.583 \\
Otter Classifier (N+C)  & 0.580 & \textbf{0.816} & \textbf{0.577} & 0.147 & \textbf{0.864} &	0.639 &	\textbf{0.641} \\ 
\hline
Otter DistMult (N+C+R)   & 0.579	& 0.810	& 0.572	& 0.145	& 0.862	& 0.629	& 0.625 \\
Otter TransE (N+C+R)   & 0.580	& 0.811 &	0.576 &	0.073 &	0.859	& 0.627	& 0.594 \\
Otter Classifier (N+C+R) & \textbf{0.582} & 0.812 &	0.574 &	0.124	& 0.860	& 0.619 & 0.600 \\ 
\hline
\end{tabular}
\caption{\label{table:universality_ubc}Information flow control and noisy links results for Otter-UBC for different scoring functions.  The table results should be compared with the results in Table \ref{tab:leaderboard} (Otter-UBC). The evaluation metrics is Pearson correlation (higher is better). N (noisy links); C (no flow control); R (regression).}
\end{table}

\subsection{Results and discussion}
\label{sec:results}

\paragraph{Knowledge enhanced representation versus vanilla representation} In Table \ref{tab:leaderboard}, we present the outcomes of the \emph{Otter-Knowledge} models, which were pretrained on graphs generated from Uniprot (U), BindingDB (B), ChemBL (C), DUDe, PrimeKG and STITCH with three different training objectives: TransE, DistMult, and binary classifier respectively. 
Also, as described in Section \ref{sec:pretraining}, we control the information propagated to the Drug/Protein entities, and manually handpick a subset of links from each database that are relevant to drug discovery. 
In all these methods, we started with the initial embeddings of sequences using ESM-1b models, while Morgan fingerprints were utilized for SMILES. We call this baseline method vanilla representation as opposed to methods utilizing knowledge-enhanced representations. The embeddings were then fine-tuned with knowledge graphs. Our results demonstrate that \emph{Otter-Knowledge} outperforms the baseline vanilla representation that lacks the enhanced knowledge from the graphs. Notably, a significant improvement was observed when we created an ensemble of 12 models trained on Otter-UBC, Otter-DUDe, Otter-PrimeKG and Otter-STITCH. We achieved state-of-the-art results on the leaderboard of the DTI DG dataset (Table \ref{tab:leaderboard}). However, for the KIBA dataset with drug split, the improvements were not substantial. As indicated in Table \ref{table:universality_ubc}, the KIBA dataset consists of only 68 drugs. The limited number of drugs makes this specific split particularly challenging, and we consider it an open challenge for future research.

\paragraph{Information flow and noisy links} Given that drug discovery predominantly involves novel entities with significantly fewer known attributes that those known during pretraining, it is important to assess if the pretraining approach can exhibit robust generalization despite the discrepancy in available data between pretraining and fine-tuning stages.
Table \ref{table:universality_ubc} shows the results of \emph{Otter-Knowledge} for Otter-UBC when (i) we do \emph{not} control the information that is propagated to Drug/Protein entities ("C" in Table \ref{table:universality_ubc}), (ii) we do \emph{not} cherry-pick a subset of links from each database that are relevant to the downstream task ("N" in Table \ref{table:universality_ubc}), (iii) regression for numerical data properties is added to the objective in addition to link prediction ("R" in Table \ref{table:universality_ubc}). Observe from the table that the results are similar to the results in Table \ref{tab:leaderboard}, with minor variations across different scoring functions and datasets. Notably, Otter Classifier with noisy links (N) and no information flow control (C) achieves comparable or even better performance than when we cherry-pick links and control the flow of information  (Table \ref{tab:leaderboard}). Slight variations indicate GNN embeddings' resilience to irrelevant noisy triples for downstream tasks. Incorporating regression objectives and information flow control doesn't seem to impact generalization or significantly improve benchmark results.

\paragraph{Morgan-fingerprint versus MolFormer} The base of the \emph{Otter-Knowledge} is to leverage the pre-trained representations, before enhancing them with additional knowledge. Thus, the initial embeddings computed for the SMILES and sequences have an impact in the results. 
Table \ref{table:impact_drug_modality} shows the results of \emph{Otter-Knowledge} using a ClassifierHead, trained on Otter-UBC for 25 epochs with different drug initial representations. We can see that the MolFormer does not give superior results compared to Morgan-fingerprint, there is room for improvement regarding learned representation over simple fingerprint-based approaches for small molecules.

\begin{table}[]
\begin{tabular}{llllllll}
\hline
Datasets                   & \textbf{DTI DG} & \multicolumn{3}{c}{\textbf{DAVIS}}               & \multicolumn{3}{c}{\textbf{KIBA}}                \\ \cline{2-8} 
Splits                     & Temporal        & Random         & Target         & Drug           & Random         & Target         & Drug           \\ \hline
MolFormer & 0.547 & 	\textbf{0.811}& \textbf{0.578}	& 0.103	& 0.838	& \textbf{0.642} & 	\textbf{0.624} \\
Morgan fingerprint& \textbf{0.574} & 	0.806	& 0.573	& \textbf{0.125}	& \textbf{0.861}	& 0.631 & 	0.619 \\
\hline
\end{tabular}
\caption{Impact of different modalities for drugs on the Otter-UDB datasets}
\label{table:impact_drug_modality}
\end{table}

\section{Related work}

We review methods to learn an effective representation from proteins, molecules and their interactions. 

\paragraph{Representation learning for proteins and small molecules}

Representation learning focuses on encoding essential information about entities, such as proteins or molecules, into low-dimensional tensors. Self-supervised algorithms using language models (LMs) have achieved remarkable success in learning protein and molecule representations by training on extensive datasets of protein sequences or linear serialization of small molecules, such as SMILES.
State of the art examples of transformer-based protein language models (PLMs) are TAPE \cite{NEURIPS2019_37f65c06}, ProteinLM \cite{DBLP:journals/corr/abs-2108-07435}, ProteinBERT \cite{proteinbert}, ESM \cite{rivesetal_2021}, Prottrans \cite{DBLP:journals/corr/abs-2007-06225}, and MSA \cite{pmlr-v139-rao21a}. They are typically trained on masked reconstruction – they learn the likelihood that a particular amino acid appears in a sequence context. Because the probability that a residue will be conserved or not across related sequences is intrinsically tied to its biological role, existent PLMs can capture co-evolutionary and inter-residue contact information \cite{NEURIPS2019_37f65c06, rivesetal_2021}, and have shown impressive performance on various tasks, such as predicting protein structure \cite{lin2022languageesm2} and function \cite{NEURIPS2019_37f65c06}. Regarding small molecules, their molecular structure can be condensed into linear notations like SMILES or SELFIES. LMs have also been used to interpret these representations, e.g., MolFormer \cite{ross2022molformer}, MolBERT \cite{fabian2020molecular}, SmilesFormer \cite{owoyemi2023smilesformer} or SELFormer \cite{yuksel2023selformer}. Both Protein and molecular representations have been fine-tuned using a contrastive learning co-embedding by Conplex \cite{Singh2022.conplex} achieving good performance in Drug-Target Interaction (DTI) prediction, surpassing state of the art approaches in the TDC-DG leaderboard which evaluates of out-of-domain generalisation \cite{NEURIPS_TDC} and achieving high specificity while detecting false positive bindings in “decoy” datasets like DUDe.

\paragraph{Knowledge enhanced pre-trained language models for proteins}
LMs do not consider existent extensive knowledge, in the form of manually curated functional and structural annotations, in human-curated domain datasets and effectively leveraging all this available factual knowledge to enhance representation learning is an open challenge. Nonetheless, prior research indicates that it can improve results in downstream learning tasks.
OntoProtein \cite{zhang2022ontoprotein} fine-tuned a PLM by reconstructing masked amino acids while minimizing the embedding distance between the contextual representation of proteins and associated gene ontology (GO) functional annotations \cite{central2023gene}. For this purpose they built ProteinKG25, a KG consisting of ~600k protein sequences and nearly five million triples. Their results show that the representations obtained where useful for classification tasks such as protein-protein interaction type, protein function, and contact prediction; but underperformed in regression tasks like homology, fluorescence, and stability.
KeAP \cite{zhou2023protein} claims to explore a more granular token-level approach, where non-masked amino acids iteratively query the associated knowledge tokens to extract helpful information (from GO) for restoring masked amino acids via cross-attention. The training process is guided only by the mask token reconstruction objective, while OntoProtein uses both contrastive learning and masked modelling simultaneously. KeAP, also trained on ProteinKG25 \cite{zhang2022ontoprotein} outperforms OntoProtein on 9 dowstream tasks. 



\paragraph{Graph-based approaches for therapeutics}

Graphs are a natural way to represent molecular interactions, signalling pathways, and disease co-morbidities. They can also be used for representation learning as they allow for the distillation of high-dimensional information about a node's neighborhood into low-dimensional vector spaces. The training objective of these representations is that similar neighborhoods are embedded close to each other in the vector space. Optimised representations can then be used to train downstream models to predict properties of specific nodes (e.g., protein function), as well as, novel associations between nodes (e.g., drug-target interactions). 
An overview on graph representation learning in biomedicine can be found in \cite{li2022graph}. 
State-of-the-art approaches have shown that incorporating multiple knowledge sources improves downstream performance. For example, DTiGEMS+ \cite{DTiGEMS} formulates the prediction of DTIs as a link prediction problem in an heterogeneous graph. TxGNN \cite{huang2023zeroshot} predicts drug indications/contraindications for rare diseases using the PrimeKG heterogeneous and multimodal KG \cite{chandak2022building}. To the best of our knowledge, we are the first to demonstrate that KGs, derived from various sources combining multiple modalities and specifically tailored to augment protein and SMILES sequences for representation learning, improved DTI downstream tasks, when evaluated using discovery benchmarks \cite{NEURIPS_TDC} wherein a majority of entities in the test data are unseen in the training data.

\section{Conclusion and future work}

In this paper we present multi-modal knowledge-enhanced representation learning for proteins and molecules, and we show its relevance and benefits for some drug discovery tasks. Our study lays the foundation for future investigations in this area with the release of multimodal KGs selectively integrating data from several datasets, and pretrained models constructed with our KGs. Our models can be utilized to acquire representations specifically tailored for drug discovery applications. Furthermore, they can serve as benchmarks for comparing and evaluating against other representation learning techniques. We have also publicly released a standard evaluation framework for assessing pretrained representations for prediction of drug-target binding affinity. We extensively analyzed different representation learning methods on three established drug-target binding datasets, and found that our approach outperformed existing methods, achieving state-of-the-art results on the TDC DG dataset.

Numerous unresolved research directions remain unexplored. Firstly, the inclusion of additional modalities, such as the 3D structure of molecules or proteins, can provide valuable insights for representation learning. Secondly, a substantial challenge lies in effectively handling a vast number of datasets, where aligning them into a single multimodal KG is not a straightforward task. Developing a learning approach capable of accommodating the dynamic input schema from diverse sources is a crucial problem to address.  Finally, assessing how representations generalize across tasks, and developing robust learning methods for representation generalization across multiple tasks under changing data distributions, remain as key research goals.

\section{Limitations}
\label{sec:limitation}
Due to license restriction of some datasets we only release the multimodal KGs with non-commercial licenses and the pretrained models that were built on these KGs. The KGs do not include 3D structure of proteins/drugs which can be an interesting modalities for future work.  



\bibliographystyle{abbrv}
\bibliography{custom}

\begin{thebibliography}{10}

\bibitem{Ashburner_2000}
M.~Ashburner, C.~A. Ball, J.~A. Blake, D.~Botstein, H.~Butler, J.~M. Cherry, A.~P. Davis, K.~Dolinski, S.~S. Dwight, J.~Eppig, M.~A. Harris, D.~P. Hill, L.~Issel-Tarver, A.~Kasarskis, S.~E. Lewis, J.~C. Matese, J.~E. Richardson, M.~Ringwald, G.~M. Rubin, and G.~Sherlock.
\newblock Gene ontology: tool for the unification of biology.
\newblock {\em Nature Genetics}, May 2000.

\bibitem{proteinbert}
N.~Brandes, D.~Ofer, Y.~Peleg, N.~Rappoport, and M.~Linial.
\newblock {ProteinBERT: a universal deep-learning model of protein sequence and function}.
\newblock {\em Bioinformatics}, 38(8):2102--2110, 02 2022.

\bibitem{central2023gene}
G.~Central, S.~A. Aleksander, J.~Balhoff, S.~Carbon, J.~M. Cherry, H.~J. Drabkin, D.~Ebert, M.~Feuermann, P.~Gaudet, N.~L. Harris, et~al.
\newblock The gene ontology knowledgebase in 2023.
\newblock {\em Genetics}, 224(1), 2023.

\bibitem{chandak2022building}
P.~Chandak, K.~Huang, and M.~Zitnik.
\newblock Building a knowledge graph to enable precision medicine.
\newblock {\em Nature Scientific Data}, 2023.

\bibitem{uniprot}
T.~U. Consortium.
\newblock {UniProt: the Universal Protein Knowledgebase in 2023}.
\newblock {\em Nucleic Acids Research}, 51(D1):D523--D531, 11 2022.

\bibitem{DBLP:journals/corr/abs-2007-06225}
A.~Elnaggar, M.~Heinzinger, C.~Dallago, G.~Rehawi, Y.~Wang, L.~Jones, T.~Gibbs, T.~Feher, C.~Angerer, M.~Steinegger, D.~Bhowmik, and B.~Rost.
\newblock Prottrans: Towards cracking the language of life's code through self-supervised deep learning and high performance computing.
\newblock {\em CoRR}, abs/2007.06225, 2020.

\bibitem{fabian2020molecular}
B.~Fabian, T.~Edlich, H.~Gaspar, M.~Segler, J.~Meyers, M.~Fiscato, and M.~Ahmed.
\newblock Molecular representation learning with language models and domain-relevant auxiliary tasks.
\newblock {\em arXiv preprint arXiv:2011.13230}, 2020.

\bibitem{gas}
M.~Fey, J.~E. Lenssen, F.~Weichert, and J.~Leskovec.
\newblock Gnnautoscale: Scalable and expressive graph neural networks via historical embeddings.
\newblock In {\em International Conference on Machine Learning}, pages 3294--3304. PMLR, 2021.

\bibitem{chembl}
A.~Gaulton, L.~J. Bellis, A.~P. Bento, J.~Chambers, M.~Davies, A.~Hersey, Y.~Light, S.~McGlinchey, D.~Michalovich, B.~Al-Lazikani, and J.~P. Overington.
\newblock {ChEMBL: a large-scale bioactivity database for drug discovery}.
\newblock {\em Nucleic Acids Research}, 40(D1):D1100--D1107, 09 2011.

\bibitem{huang2023zeroshot}
K.~Huang, P.~Chandak, Q.~Wang, S.~Havaldar, A.~Vaid, J.~Leskovec, G.~Nadkarni, B.~Glicksberg, N.~Gehlenborg, and M.~Zitnik.
\newblock Zero-shot prediction of therapeutic use with geometric deep learning and clinician centered design.
\newblock {\em medRxiv}, 2023.

\bibitem{NEURIPS_TDC}
K.~Huang, T.~Fu, W.~Gao, Y.~Zhao, Y.~Roohani, J.~Leskovec, C.~Coley, C.~Xiao, J.~Sun, and M.~Zitnik.
\newblock Therapeutics data commons: Machine learning datasets and tasks for drug discovery and development.
\newblock In J.~Vanschoren and S.~Yeung, editors, {\em Proceedings of the Neural Information Processing Systems Track on Datasets and Benchmarks}, volume~1. Curran, 2021.

\bibitem{Huang2022}
K.~Huang, T.~Fu, W.~Gao, Y.~Zhao, Y.~Roohani, J.~Leskovec, C.~W. Coley, C.~Xiao, J.~Sun, and M.~Zitnik.
\newblock Artificial intelligence foundation for therapeutic science.
\newblock {\em Nature chemical biology}, 2022.

\bibitem{drugbank}
V.~Law, C.~Knox, Y.~Djoumbou, T.~Jewison, A.~C. Guo, Y.~Liu, A.~Maciejewski, D.~Arndt, M.~Wilson, V.~Neveu, A.~Tang, G.~Gabriel, C.~Ly, S.~Adamjee, Z.~T. Dame, B.~Han, Y.~Zhou, and D.~S. Wishart.
\newblock {DrugBank 4.0: shedding new light on drug metabolism}.
\newblock {\em Nucleic Acids Research}, 42(D1):D1091--D1097, 11 2013.

\bibitem{li2022graph}
M.~M. Li, K.~Huang, and M.~Zitnik.
\newblock Graph representation learning in biomedicine and healthcare.
\newblock {\em Nature Biomedical Engineering}, pages 1--17, 2022.

\bibitem{lin2022languageesm2}
Z.~Lin, H.~Akin, R.~Rao, B.~Hie, Z.~Zhu, W.~Lu, N.~Smetanin, A.~dos Santos~Costa, M.~Fazel-Zarandi, T.~Sercu, S.~Candido, et~al.
\newblock Language models of protein sequences at the scale of evolution enable accurate structure prediction.
\newblock {\em bioRxiv}, 2022.

\bibitem{liu2007bindingdb}
T.~Liu, Y.~Lin, X.~Wen, R.~N. Jorissen, and M.~K. Gilson.
\newblock Bindingdb: a web-accessible database of experimentally determined protein--ligand binding affinities.
\newblock {\em Nucleic acids research}, 35(suppl\_1):D198--D201, 2007.

\bibitem{dude}
M.~M. Mysinger, M.~Carchia, J.~J. Irwin, and B.~K. Shoichet.
\newblock Directory of useful decoys, enhanced (dud-e): better ligands and decoys for better benchmarking.
\newblock {\em Journal of medicinal chemistry}, 55(14):6582--6594, 2012.

\bibitem{opentargets}
D.~Ochoa, A.~Hercules, M.~Carmona, D.~Suveges, J.~Baker, C.~Malangone, I.~Lopez, A.~Miranda, C.~Cruz-Castillo, L.~Fumis, M.~Bernal-Llinares, K.~Tsukanov, H.~Cornu, K.~Tsirigos, O.~Razuvayevskaya, A.~Buniello, J.~Schwartzentruber, M.~Karim, B.~Ariano, R.~Martinez Osorio, J.~Ferrer, X.~Ge, S.~Machlitt-Northen, A.~Gonzalez-Uriarte, S.~Saha, S.~Tirunagari, C.~Mehta, J.~Roldán-Romero, S.~Horswell, S.~Young, M.~Ghoussaini, D.~Hulcoop, I.~Dunham, and E.~McDonagh.
\newblock {The next-generation Open Targets Platform: reimagined, redesigned, rebuilt}.
\newblock {\em Nucleic Acids Research}, 51(D1):D1353--D1359, 11 2022.

\bibitem{open-targets}
D.~Ochoa, A.~Hercules, M.~Carmona, D.~Suveges, J.~Baker, C.~Malangone, I.~Lopez, A.~Miranda, C.~Cruz-Castillo, L.~Fumis, M.~Bernal-Llinares, K.~Tsukanov, H.~Cornu, K.~Tsirigos, O.~Razuvayevskaya, A.~Buniello, J.~Schwartzentruber, M.~Karim, B.~Ariano, R.~Martinez Osorio, J.~Ferrer, X.~Ge, S.~Machlitt-Northen, A.~Gonzalez-Uriarte, S.~Saha, S.~Tirunagari, C.~Mehta, J.~Roldán-Romero, S.~Horswell, S.~Young, M.~Ghoussaini, D.~Hulcoop, I.~Dunham, and E.~McDonagh.
\newblock {The next-generation Open Targets Platform: reimagined, redesigned, rebuilt}.
\newblock {\em Nucleic Acids Research}, 51(D1):D1353--D1359, 11 2022.

\bibitem{owoyemi2023smilesformer}
J.~Owoyemi and N.~Medzhidov.
\newblock Smilesformer: Language model for molecular design.
\newblock 2023.

\bibitem{deepdta}
H.~{\"O}zt{\"u}rk, A.~{\"O}zg{\"u}r, and E.~Ozkirimli.
\newblock Deepdta: deep drug--target binding affinity prediction.
\newblock {\em Bioinformatics}, 34(17):i821--i829, 2018.

\bibitem{NEURIPS2019_37f65c06}
R.~Rao, N.~Bhattacharya, N.~Thomas, Y.~Duan, P.~Chen, J.~Canny, P.~Abbeel, and Y.~Song.
\newblock Evaluating protein transfer learning with tape.
\newblock In H.~Wallach, H.~Larochelle, A.~Beygelzimer, F.~d\textquotesingle Alch\'{e}-Buc, E.~Fox, and R.~Garnett, editors, {\em Advances in Neural Information Processing Systems}, volume~32. Curran Associates, Inc., 2019.

\bibitem{pmlr-v139-rao21a}
R.~M. Rao, J.~Liu, R.~Verkuil, J.~Meier, J.~Canny, P.~Abbeel, T.~Sercu, and A.~Rives.
\newblock Msa transformer.
\newblock In M.~Meila and T.~Zhang, editors, {\em Proceedings of the 38th International Conference on Machine Learning}, volume 139 of {\em Proceedings of Machine Learning Research}, pages 8844--8856. PMLR, 18--24 Jul 2021.

\bibitem{reimers-2019-sentence-bert}
N.~Reimers and I.~Gurevych.
\newblock Sentence-bert: Sentence embeddings using siamese bert-networks.
\newblock In {\em Proceedings of the 2019 Conference on Empirical Methods in Natural Language Processing}. Association for Computational Linguistics, 11 2019.

\bibitem{rivesetal_2021}
A.~Rives, J.~Meier, T.~Sercu, S.~Goyal, Z.~Lin, J.~Liu, D.~Guo, M.~Ott, C.~Zitnick, J.~Ma, and R.~Fergus.
\newblock Biological structure and function emerge from scaling unsupervised learning to 250 million protein sequences.
\newblock {\em Proceedings of the National Academy of Sciences of the United States of America}, 118(15), Apr. 2021.

\bibitem{rives2019biological}
A.~Rives, J.~Meier, T.~Sercu, S.~Goyal, Z.~Lin, J.~Liu, D.~Guo, M.~Ott, C.~L. Zitnick, J.~Ma, and R.~Fergus.
\newblock Biological structure and function emerge from scaling unsupervised learning to 250 million protein sequences.
\newblock {\em PNAS}, 2019.

\bibitem{molformer}
J.~Ross, B.~Belgodere, V.~Chenthamarakshan, I.~Padhi, Y.~Mroueh, and P.~Das.
\newblock {Large-scale chemical language representations capture molecular structure and properties}.
\newblock {\em Nature Machine Intelligence}, 4(12):1256--1264, 2022.

\bibitem{ross2022molformer}
J.~Ross, B.~Belgodere, V.~Chenthamarakshan, I.~Padhi, Y.~Mroueh, and P.~Das.
\newblock Molformer: Large scale chemical language representations capture molecular structure and properties.
\newblock 2022.

\bibitem{Schlichtkrull2018Modeling}
M.~Schlichtkrull, T.~N. Kipf, P.~Bloem, R.~van den Berg, I.~Titov, and M.~Welling.
\newblock Modeling relational data with graph convolutional networks.
\newblock In A.~Gangemi, R.~Navigli, M.-E. Vidal, P.~Hitzler, R.~Troncy, L.~Hollink, A.~Tordai, and M.~Alam, editors, {\em The Semantic Web}, pages 593--607, Cham, 2018. Springer International Publishing.

\bibitem{Singh2022.conplex}
R.~Singh, S.~Sledzieski, L.~Cowen, and B.~Berger.
\newblock Learning the drug-target interaction lexicon.
\newblock {\em bioRxiv}, 2022.

\bibitem{sledzieski2022adapting}
S.~Sledzieski, R.~Singh, L.~Cowen, and B.~Berger.
\newblock Adapting protein language models for rapid dti prediction.
\newblock {\em bioRxiv}, pages 2022--11, 2022.

\bibitem{https://doi.org/10.1002/ail2.20}
P.~W.~J. Staar, M.~Dolfi, and C.~Auer.
\newblock Corpus processing service: A knowledge graph platform to perform deep data exploration on corpora.
\newblock {\em Applied AI Letters}, 1(2):e20, 2020.

\bibitem{szklarczyk2023steing}
D.~Szklarczyk, R.~Kirsch, M.~Koutrouli, K.~Nastou, F.~Mehryary, R.~Hachilif, A.~Gable, T.~Fang, N.~Doncheva, S.~Pyysalo, P.~Bork, L.~J. Jensen, and C.~von Mering.
\newblock The string database in 2023: protein-protein association networks and functional enrichment analyses for any sequenced genome of interest.
\newblock {\em Nucleic acids research}, 51(D1):D638--D646, 2023.

\bibitem{szklarczyk2016stitch}
D.~Szklarczyk, A.~Santos, C.~von Mering, L.~J. Jensen, P.~Bork, and M.~Kuhn.
\newblock Stitch 5: augmenting protein-chemical interaction networks with tissue and affinity data.
\newblock {\em Nucleic acids research}, 44(D1):D380--D384, 2016.

\bibitem{DTiGEMS}
M.~Thafar, R.~Olayan, H.~Ashoor, S.~Albaradei, V.~Bajic, X.~Gao, T.~Gojobori, and M.~Essack.
\newblock Dtigems+: drug–target interaction prediction using graph embedding, graph mining, and similarity-based techniques.
\newblock {\em Journal of Cheminformatics}, 12(1), July 2020.
\newblock KAUST Repository Item: Exported on 2020-10-01 Acknowledged KAUST grant number(s): BAS/1/1606-01-01, BAS/1/1059-01-01, BAS/1/1624-01-01, FCC/1/1976-17-01, FCC/1/1976-26-01. Acknowledgements: The research reported in this publication was supported by the King Abdullah University of Science and Technology (KAUST).

\bibitem{DBLP:journals/corr/abs-2108-07435}
Y.~Xiao, J.~Qiu, Z.~Li, C.~Hsieh, and J.~Tang.
\newblock Modeling protein using large-scale pretrain language model.
\newblock {\em CoRR}, abs/2108.07435, 2021.

\bibitem{yuksel2023selformer}
A.~Y{\"u}ksel, E.~Ulusoy, A.~{\"U}nl{\"u}, G.~Deniz, and T.~Do{\u{g}}an.
\newblock Selformer: Molecular representation learning via selfies language models.
\newblock {\em arXiv preprint arXiv:2304.04662}, 2023.

\bibitem{zhang2022ontoprotein}
N.~Zhang, Z.~Bi, X.~Liang, S.~Cheng, H.~Hong, S.~Deng, Q.~Zhang, J.~Lian, and H.~Chen.
\newblock Ontoprotein: Protein pretraining with gene ontology embedding.
\newblock In {\em International Conference on Learning Representations}, 2022.

\bibitem{zhou2023protein}
H.-Y. Zhou, Y.~Fu, Z.~Zhang, B.~Cheng, and Y.~Yu.
\newblock Protein representation learning via knowledge enhanced primary structure reasoning.
\newblock In {\em The Eleventh International Conference on Learning Representations}, 2023.

\bibitem{ZHOU2020graph}
J.~Zhou, G.~Cui, S.~Hu, Z.~Zhang, C.~Yang, Z.~Liu, L.~Wang, C.~Li, and M.~Sun.
\newblock Graph neural networks: A review of methods and applications.
\newblock {\em AI Open}, 1:57--81, 2020.

\end{thebibliography}

\newpage

\newpage
\appendix

\section{Appendix}

\subsection{Experimental settings}

\begin{table}[h]
\centering
\begin{tabular}{ll|ll}
\hline
\multicolumn{2}{c|}{\textbf{GNN models}}   & \multicolumn{2}{c}{\textbf{Downstream models}}           \\ \hline
\textbf{Hyperparameters} & \textbf{Values} & \textbf{Hyperparameters}      & \textbf{Values}          \\ \hline
\# graph partitions      & 100             & Learning rate                 & 5e-4 (5e-5 for TDC DG)   \\
Learning rate            & 1e-5            & \# training steps             & 10000 (20000 for TDC DG) \\
\# layers                & 2               & Batch size                    & 256                      \\
Projection size          & 64              & \# layers                     & 2                        \\
Hidden size              & 128             & Initial embedding hidden size & {[}1024, 512{]}          \\
Output size              & 128             & GNN embedding hidden size     & {[}1024, 1024{]}         \\
Train/val                & 0.9             & Activation                    & RELU                     \\
Training epochs          & 35              & Random seeds                  & 0-5                      \\
Acivation                & RELU            & \# runs                       & 6                        \\
Random seeds             & 3000            &                               &                          \\ 
Max train time             & 24 hours            &                               &                          \\
\hline
\end{tabular}
\caption{Hyperparameters setting for pretrained GNN models and downstream models \label{tab:hyperparameters}}
\end{table}

The downstream models comprise of two parallel networks. In the first one, the initial embeddings of drugs and proteins, obtained using the Morgan-fingerprint and ESM-1b models, are concatenated. These embeddings undergo a transformation through two linear layers and RELU activation before predicting the binding affinity. The second architecture involves the concatenation of GNN embeddings of drugs and proteins. Similarly, these embeddings are transformed using two linear layers and RELU activation, followed by predicting the binding affinity.

To obtain the final prediction for the binding affinity, the outputs of the two parallel networks are summed. Initially, a simpler approach was attempted, which involved concatenating all initial embeddings and GNN embeddings of drugs and proteins, followed by transforming these embeddings using a single network. However, this approach did not yield significant results due to the correlation between GNN embeddings and the initial embeddings of drugs and proteins. All the hyperparameters for the downstream models are provided in Table \ref{tab:hyperparameters}.

Since the primary focus of the work is the demonstration of the value of enriched representation using diverse information sources, rather than the discovery of optimal neural network structures, we did not perform an automated search for architectures. Instead, we established the size of the GNN (Graph Neural Network) and downstream models as specified in Table \ref{tab:hyperparameters}. To ensure reliable results, we conducted multiple downstream experiments using six different random seeds ranging from 0 to 5, following the recommended approach outlined in the TDC work \cite{NEURIPS_TDC}. The reported figures in this paper are the average outcomes obtained from these experiments.

The results presented in Table \ref{tab:leaderboard} rely on pretrained models that have restricted the information they can access for Protein/Drug entities. These entities are only allowed to utilize protein sequences and smiles.  Inference tasks follow the pretraining tasks to acquire pretrained GNN embeddings for drugs/proteins in the downstream tasks. Even so Table \ref{table:universality_ubc} shows that information flow controls to drugs/proteins is not necessary to achieve good performance on the downstream tasks.  

There are two sets of hyper-parameters, one concerning GNN models training and the other concerning donwstream model training. The learning rate of the first set of hyper-parameters was chosen in the range 0.01-0.00001 (log scale) to optimize the loss of the validation split of the pretraining tasks.

\begin{figure}[h]
\centering
\includegraphics[width=\textwidth]{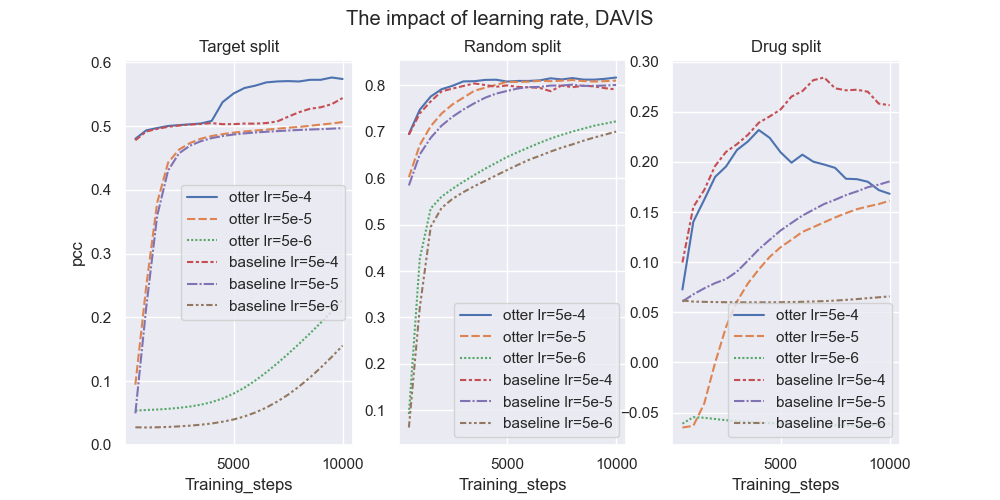}
\caption{Learning rate impact on downstream DAVIS tasks \label{fig:davis learning rate}}
\end{figure}

\begin{figure}[h]
\centering
\includegraphics[width=\textwidth]{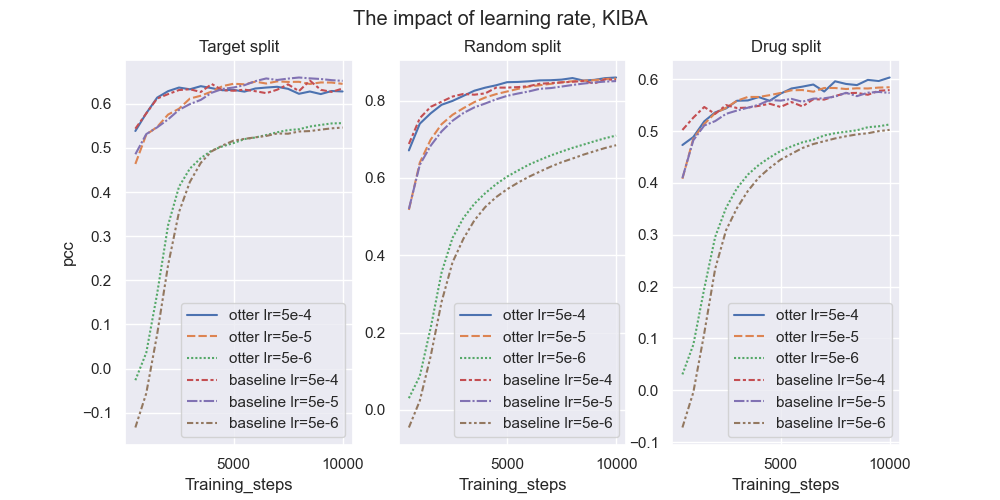}
\caption{Learning rate impact on downstream  KIBA tasks \label{fig:kiba learning rate}}
\end{figure}

The learning rate of the second set of hyper-parameters for the downstream tasks were chosen in the range 0.0005, 0.00005, 0.000005 based on the Pearson correlation score observed on the validation dataset of the TDC-DG dataset. For DAVIS and KIBA where the data is small (has no validation split), we have discussed how to choose the learning rate in Figure \mbox{\ref{fig:davis learning rate}} and Figure \mbox{\ref{fig:kiba learning rate}}. The performance of the baseline and the proposed approach in the paper varied with different values of learning rate, but we always observed better results with the proposed approach in this paper compared to the baseline without knowledge enhanced representation.



\subsection{Model release}
We made available a total of 12 pretrained models that were utilized in our experiments, as indicated in Table \ref{tab:leaderboard}. Additionally, we have provided the source code necessary for performing inference using these models. To utilize these pretrained models, users can utilize the inference API and provide new protein sequences or new smiles as input. The API will then return both the initial embeddings and the pretrained GNN embeddings for the given input. 

We made available a total of 12 pretrained models that were utilized in our experiments, as indicated in Table \ref{tab:leaderboard}. Additionally, we have provided the source code necessary for performing inference using these models. To utilize these pretrained models, users can utilize the inference API and provide new protein sequences or new smiles as input. The API will then return both the initial embeddings and the pretrained GNN embeddings for the given input. 

Table \ref{tab:models} provides a summary of the models. Each model's name reflects the dataset it was trained on and the scoring function employed during pretraining. For instance, the model otter\_ubc\_transe denotes the one trained on the Uniprot, BindingDB, and ChemBL (Otter-UBC) dataset using the TransE scoring function. All models underwent training for either 35 epochs or within a maximum time frame of 24 hours.

The models can be found in the Huggingface hub \footnote{\url{https://huggingface.co/models?sort=downloads&search=ibm/otter_}}. The name of the models in Table \ref{tab:models} is also the name of the models in the hub, under the IBM organization space. For example the otter\_ubc\_transe model could be found in the hub in \url{https://huggingface.co/ibm/otter_ubc_transe}.

\begin{table}[]
\centering
\begin{tabular}{lllll}
\hline
\textbf{Models}     & \textbf{Pretrain datasets}      & \textbf{Learning objective} &\textbf{License} \\ \hline
otter\_ubc\_transe         & Otter-UBC & TransE                       & MIT              \\
otter\_ubc\_distmult       & Otter-UBC & DistMult                    & MIT              \\
otter\_ubc\_classifier     & Otter-UBC & Classifier                  & MIT              \\
otter\_dude\_transe        & Otter-DUDe                               & TransE                      & MIT              \\
otter\_dude\_distmult      & Otter-DUDe                               & DistMult                    & MIT              \\
otter\_dude\_classifier    & Otter-DUDe                               & Classifier                  & MIT              \\
otter\_primekg\_transe     & Otter-PrimeKG                            & TransE                      & MIT              \\
otter\_primekg\_distmult   & Otter-PrimeKG                            & DistMult                    & MIT              \\
otter\_primekg\_classifier & Otter-PrimeKG                            & Classifier                  & MIT              \\
otter\_stitch\_transe      & Otter-Stitch                             & TransE                      & MIT              \\
otter\_stitch\_distmult    & Otter-Stitch                             & DistMult                    & MIT              \\
otter\_stitch\_classifier  & Otter-Stitch                             & Classifier                  & MIT              \\ \hline
\end{tabular}
\caption{\label{tab:models} Summary of the released pretrained models.}
\end{table}

\subsection{Datasets release}

\begin{figure}[ht]
\centering
\includegraphics[width=\textwidth]{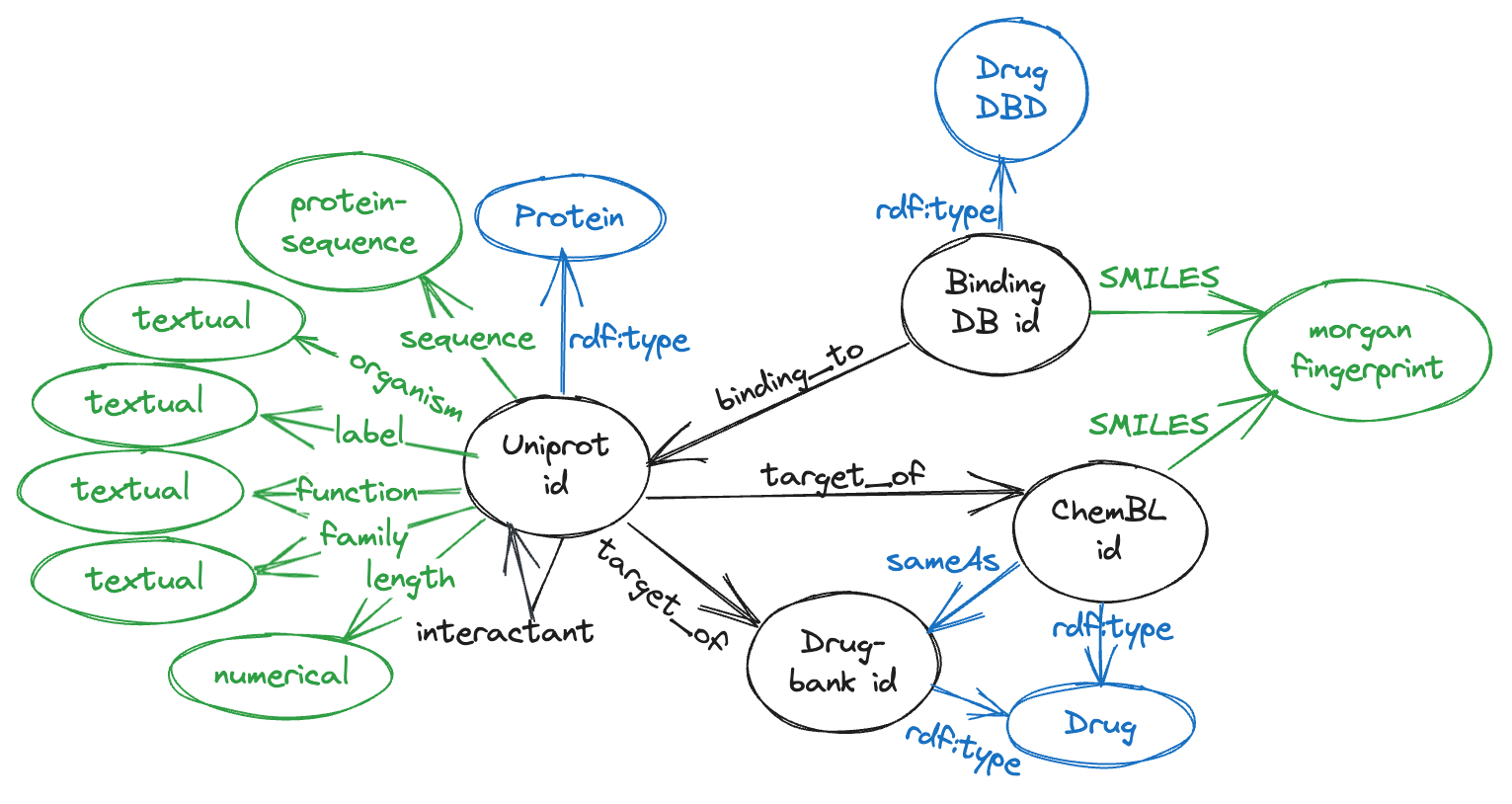}
\caption{\label{fig:workflow-schema} Schema for Otter-UBC Knowledge Graph.}
\end{figure}

\begin{figure}[t]
\centering
\includegraphics[width=\textwidth]{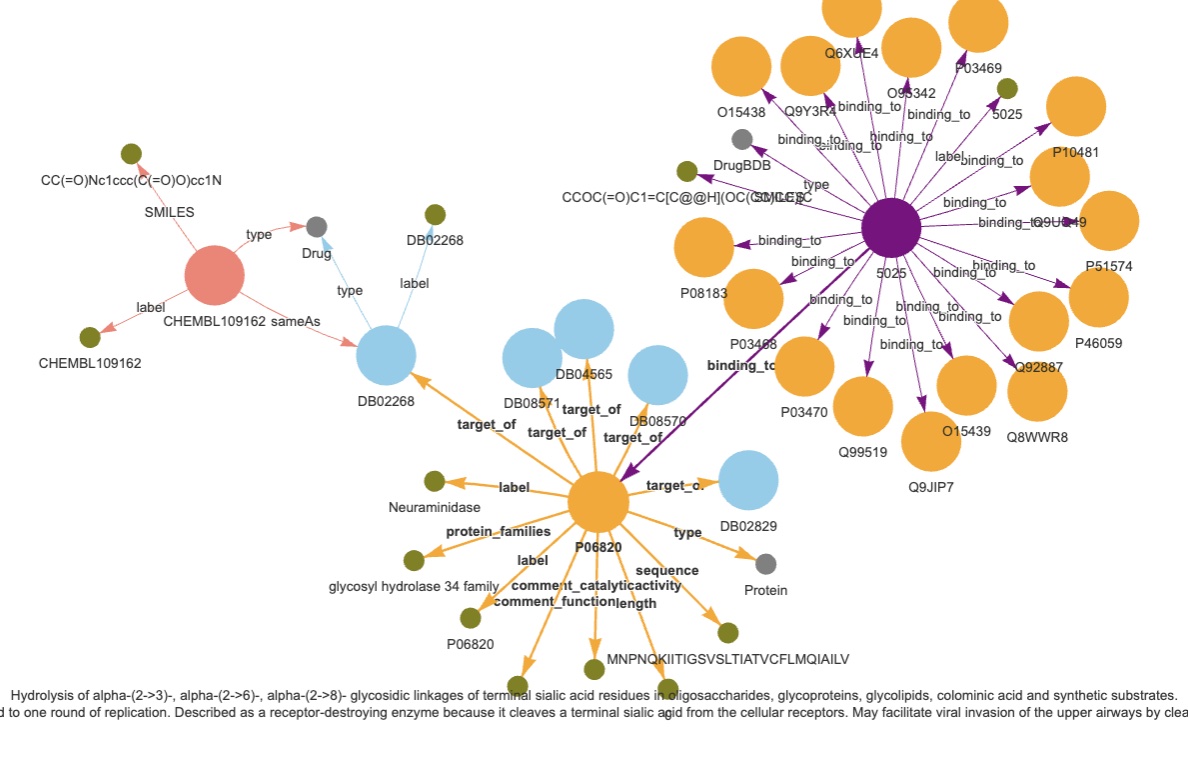}
\caption{\label{fig:kg_example} Subset of a KG  in Otter-UBC. Attribute nodes are coloured in green, type classes in blue, uniprot, chembl, drugbank and bindingDB entities are coloured in orange, salmon, blue and purple respectively}
\end{figure}

We release the 4 KGs used to create the 12 pretrained models released. These KGs were constructed from the original dataset sources using a JSON schema that specifies the type of entities and their respective namespaces. We use namespaces to assign a unique identifier to each entity based on its source, so that in the KGs we can connect and cross reference entities within and across dataset sources (an exception is PrimeKG, which requires custom processing of nodes because of the format of the original CSV files - we use however the same relations as defined by PrimeKG). For each entity type, the schema defines the relevant object and data properties and the modalities for both entities and attributes. 

Duplicates are therefore handled effectively through the user provided schema by specifying the namespace of the original source and the unique entity id (e.g, if an entity in BindingDB references an entity with a Uniprot ID, the schema needs to specify that the target entity belongs to the Uniprot namespace). Additionally, an owl:sameAs relation can be specified in the schema to leverage the links explicitly established across entities with different IDs across data sources. Nonetheless, coping with the intricacies of automatic alignment for entities not explicitly link in the original sources, along with relations that may be semantically similar yet have different names across datasets, constitutes a research challenge. The down effect of this is that the model may face difficulties to learn to transfer the commonalities between two (duplicated) entities or relations, if there isn't an explicit mechanism to force the model to do so. Designing a method that enables the model to learn to transfer across sources with varying schemas remains an open research question.

Figure \ref{fig:workflow-schema} shows an example schema for the Otter-UBC KG, which comprises 9,386,418 RDF triples obtained by combining factual information from Uniprot, BindingDB and a subset of ChemBL Drugs molecules obtained from the Open Targets platform \cite{open-targets}; however for training the GNN we omitted triples with the \texttt{rdf:type} predicate, as this information is implicit in the modality and it would result in a very densely connected graph (e.g, when considering all nodes of type \emph{Protein}), resulting in 6,207,654 triples. 
 
Figure \ref{fig:kg_example} shows a portion of the resulting KG where an entity of type \texttt{Protein} defined with its UniProt ID and namedspace (\texttt{P06820}) has data attributes such as sequence, label (\texttt{Neuraminidase}), family (\texttt{glycosyl hydrolase 34 family}), as well as object properties from and to other entities (e.g, a \texttt{target\_of} relation to drug \texttt{DB02268}, which is also the object of a \texttt{owl:sameAs} relation from \texttt{CHEMBL1091644} ).

We note that for STITCH the raw data was additionally processed to filter out interactions with the highest likelihood (greater than 0.9) and to remove duplicated interactions based on chemical SMILES and protein sequence. Also, files containing information about interactions, SMILES and protein sequences were joined into a single CSV file for which schema was designed for KG creation.

The datasets are serialised in N-Triples\footnote{\url{https://www.w3.org/TR/n-triples/}}, a  line-based, plain text serialisation format for RDF triples and the pretrained models are accessible at the link \footnote{\url{https://github.com/IBM/otter-knowledge}}. The original tabular datasets were either downloaded from their corresponding websites or accessed using Deep Search\footnote{\url{https://ds4sd.github.io}}, such as in the case of Uniprot SwissProt. Leveraging state-of-the-art AI methods, Deep Search collects, converts, enriches, and links large document collections \cite{https://doi.org/10.1002/ail2.20} and it comes pre-loaded with millions of standard technical documents from many public data sources.

The same way as the models, the datasets have been released and can be found in the Huggingface hub. The names of the datasets in the hub, under the IBM organization space, are the next ones: 
\begin{itemize}
    \item otter\_uniprot\_bindingdb: MKG comprising entities and relations from Uniprot and BindingDB. 
    \item otter\_uniprot\_bindingdb\_chembl: (Otter-UBC) MKG comprising entities and relations from Uniprot, BindingDB and ChemBL. 
    \item otter\_dude: (Otter-DUDe) MKG with data from the preprocessed version of DUDe
    \item otter\_primekg: (Otter-PrimeKG) MKG with data from the PrimeKG
    \item otter\_stitch: (Otter-STITCH) MKG with data from the preprocessed version of STITCH
\end{itemize}

Note that when training the GNN, the data attributes of a Protein or Drug entity node are treated as graph nodes with their own modalities (e.g textual, numerical, sequence, smiles), rather than considering them just as features or datatype attributes of the entity node. Therefore, the entity node (e.g, representing a Protein or Drug) receives information from the various available modalities through convolutional message passing in each iteration. The advantage is that entity nodes are not required to carry all multimodal properties or project large property vectors with missing values. Instead, the projection is done per modality and only when such a modality exists for the entity.

\subsection{Information control flow and Restricted (not noisy) link prediction}
The graph might be large, with many links that may not be relevant for the tasks ahead. In the process of GNN pretraining , substantial amounts of known attributes are available for both proteins and molecules in the KGs. However, when it comes to fine-tuning for specific tasks like predicting drug-target binding affinity, this information is severely limited for novel proteins and drugs. In such cases, only basic details such as protein amino acid sequences and molecule SMILES are accessible. We aim to study if this discrepancy in data between pretraining and fine-tuning phases could lead to a decrease in the accuracy of binding affinity prediction for newly introduced entities. 

To assess the impact of i) regulating the information propagation / flow and ii) restricted link predictions, we have a list of permissible specific links in each KG that matter for predicting drug-target binding in downstream tasks. Our tests showed results for both cases: when we control the information flow and predict links only from our picks, and another where we considered all links regardless of the task. 

In the information control flow process within each iteration of GNN training, exclusively the messages originating from approved links are employed to update node embedding. As illustration, consider a protein node; the accepted link types to control the information flow include sequence attributes. Similarly, for a molecule node, permissible link types encompass the SMILES attributes. This selection stems from the fact that, during the testing phase, newly introduced Proteins and Molecules lack attributes beyond sequence and SMILES.

 For the experiments with restricted links prediction, these are the ones we chose for each knowledge graph:
\begin{itemize}
    \item Otter-UBC: target of, binding to
    \item Otter-DUDe: binding to
    \item Otter-PrimeKG: protein protein interaction, drug drug synergistic-interaction, drug protein transporter, drug protein target, drug protein enzyme
    \item Otter-STITCH: binding to
\end{itemize}

The experimental results without control flow and  link predictions with all possible links, including potentially noisy links to the downstream tasks, are demonstrated in Table 4. The results prove our method works well even with noisy links, as the outcomes are comparable in both cases – whether we limit the links or not. Additionally, using extra modalities in pre-training do not impair generalisation to fine-tuning.

\subsection{Results with Otter-UDB and Otter-UB}

The outcomes of the model that underwent pretraining on Uniprot and BindingDB (Otter-UB) compared to the model that underwent pretraining on Uniprot, Drugbank, and BindingDB (Otter-UDB) are presented in Table \ref{tab:udb vs ub }. Although we are unable to share the processed Drugbank datasets because of licensing limitations, it is evident that incorporating Drugbank does not lead to significantly different results compared to using UB alone.

\begin{table}[]
\small
\begin{tabular}{lllllllll}
\hline
                            & \textbf{Downstream}                                                                                                   & \textbf{DTI DG}            & \multicolumn{3}{c}{\textbf{DAVIS}}                                                                                       & \multicolumn{3}{c}{\textbf{KIBA}}                                                                   \\ \cline{3-9} 
\textbf{Upstream} & Splits                                                                                                                         & \multicolumn{1}{l|}{Temporal}                        & Random                          & Target                          & \multicolumn{1}{l|}{Drug}                            & Random                          & Target                          & Drug                            \\ \hline
\multirow{3}{*}{Otter-UDB}        & Otter DistMult     & \multicolumn{1}{l|}{0.577}  & 0.809    & 0.571   & 
                            \multicolumn{1}{l|}{0.144}                         & 0.858    & 0.636   & 0.587                           \\
                            & Otter TransE       & \multicolumn{1}{l|}{0.581}    & 0.805  & 0.571   & 
                            \multicolumn{1}{l|}{0.132}                           & 0.857  & 0.633   & 0.585                           \\
                            & Otter Classifier    & \multicolumn{1}{l|}{0.577}   & 0.811  & 0.574   &
                            \multicolumn{1}{l|}{0.129}                           & 0.861  & 0.627   & 0.619                           \\ \hline
\multirow{3}{*}{Otter-UB}         & Otter DistMult      & \multicolumn{1}{l|}{0.576}   & 0.808  & 0.574   &  
                            \multicolumn{1}{l|}{0.126}                           & 0.858  & 0.627   & 0.596                               \\
                            & Otter TransE         & \multicolumn{1}{l|}{0.579}  & 0.808  & 0.571   & 
                            \multicolumn{1}{l|}{0.162}                           & 0.858  & 0.633   & 0.581
                                    \\
                            & Otter Classifier     & \multicolumn{1}{l|}{0.577}  & 0.809  & 0.575   & 
                            \multicolumn{1}{l|}{0.138}                           & 0.860  & 0.634   &  0.617     \\            \hline
\end{tabular}
\caption{\label{tab:udb vs ub } Results of knowledge enhanced representation on three standard drug-target binding affinity prediction benchmarks datasets with different splits of the models pretrained with Otter-UDB and Otter-UB respectively.}
\end{table}


\begin{table}[]
\begin{tabular}{llllllll}
\hline
Datasets (Otter-UDB)                  & \textbf{DTI DG} & \multicolumn{3}{c}{\textbf{DAVIS}}               & \multicolumn{3}{c}{\textbf{KIBA}}                \\ \cline{2-8} 
Splits                     & Temporal        & Random         & Target         & Drug           & Random         & Target         & Drug           \\ \hline

Otter DistMult (U)  & 0.576           & \textbf{0.811}          & \textbf{0.574}          & 0.143          & 0.857          & \textbf{0.642} & 0.580          \\
Otter TransE (U)    & 0.576  & 0.807          & 0.573          & \textbf{0.147}          & 0.857          & 0.641          & 0.581          \\
Otter Classifier (U) & \textbf{0.577}           & 0.808 & 0.573 & 0.136          & 0.858 & 0.639          & 0.592 \\ 
\hline
Otter DistMult (U+R)   & 0.576         & 0.809  & 0.573          & 0.141                    & \textbf{0.859}          & 0.635 & \textbf{0.597}          \\
Otter TransE  (U+R)    & 0.575  & 0.808          & \textbf{0.574}          & 0.145          & 0.858          & 0.634          & 0.596          \\
Otter Classifier (U+R) & 0.576           & 0.808 & 0.573 & 0.136          & 0.858 & 0.639          & 0.592 \\ 
\hline
DistMult (N) & 0.574           & 0.807          & 0.571          & 0.128         & 0.861          & {0.633} & 0.597          \\
TransE (N) & 0.573  & \textbf{0.812}          & \textbf{0.575}          & \textbf{0.178}          & \textbf{0.862}          & 0.633          & 0.596          \\
Classifier (N) & \textbf{0.579}           & {0.808} & \textbf{0.575} & 0.141          & {0.860} & 0.638          & \textbf{0.603} \\ 
\hline
\end{tabular}
\caption{Universality results for Otter-UDB, where the scoring function considers all the knowledge graph links during the training. The table results should be compared with the results in Table \ref{tab:leaderboard} (Otter-UDB). The evaluation metrics is Pearson correlation (higher is better).}
\label{table:universality}
\end{table}

%
%

\subsection{Computing resource and training time}
Each pretrained model was trained using a single A100 GPU. The training process was restricted to a maximum duration of 24 hours and a maximum of 35 epochs for each pretraining graph. The specific number of epochs within the 24-hour time frame could differ based on the size of the input graph.

For managing the large graphs like the Otter-UBC, we utilized 200GB of CPU memory. The storage of the initial embeddings for the modality required the highest amount of memory compared to other components. With the implementation of the GAS methods explained in Section \ref{sec:pretraining}, the GNN training process was able to handle the large graphs effectively, even when constrained by GPU memory limitations.

\subsection{Downstream task open source}
As  we have discussed above, the neural architectures in the TDC framework  are not optimal for the TDC DG benchmarks because they are designed to learn proteins/drugs representation from scratch. For pretrained  proteins/drugs representation evaluation we created a new architecture that provide better results for both the baseline and the GNN representation. We made these architectures available for reproducible results. Additionally, we have open-sourced the source code of the ensemble methods used in our work.

\subsection{Results with Spearman rank correlation}
In this section, we present supplementary experiments involving the Spearman rank correlation, an alternative metric to Pearson's correlation.

The outcomes presented in Table \ref{table:spearman} were collected from the model pretraining on UBC knowledge graphs. Evidently, the augmented representation utilizing data from UBC knowledge graphs outperforms the outcomes obtained from utilizing basic pretrained embeddings from ESM models for proteins and the Morgan Fingerprint for drugs. These findings underscore the robustness of the enhanced representation across diverse evaluation metrics.

\begin{table}[]
\begin{tabular}{llllllll}
\hline
Datasets                  & \textbf{DTI DG} & \multicolumn{3}{c}{\textbf{DAVIS}}               & \multicolumn{3}{c}{\textbf{KIBA}}                \\ \cline{2-8} 
Splits                     & Temporal        & Random         & Target         & Drug           & Random         & Target         & Drug           \\ \hline

\hline
Otter Distmult & 0.590 & 0.635 & 0.510	& 0.146 & 0.837	& 0.570	& 0.549 \\
Otter TransE & 0.592 & \textbf{0.638} & 0.506& 0.128 & 0.833	& \textbf{0.599}	& 0.536\\
Otter Classifier & \textbf{0.593} & 0.637 & \textbf{0.512}	& 0.102 & \textbf{0.840}	& 0.580	& \textbf{0.579} \\
ESM+Morgan Fingerprint & 0.591 & 	0.634 &	0.504 & \textbf{0.268} &	0.826 &	0.579	& 0.536 \\
\hline
\end{tabular}
\caption{\label{table:spearman} The Spearman's rank correlation of the models trained on Uniprot, BindingDB and ChemBL (UBC) and the baseline models using  ESM and Morgan Fingerprint embeddings for proteins and drugs respectively.}
\end{table}

\end{document}